\documentclass[letterpaper, 10 pt, conference]{ieeeconf}  

\IEEEoverridecommandlockouts                              

\usepackage{graphics} 
\usepackage{epsfig} 
\usepackage{mathptmx} 
\usepackage{times} 
\usepackage{amsmath} 
\usepackage{amssymb}  
\usepackage{fix2col}
\usepackage{ctable}
\usepackage{cite}

\usepackage{xcolor}
\usepackage{color, colortbl}
\usepackage{color}

\graphicspath{{Figures/}}

\definecolor{americanrose}{rgb}{1.0, 0.01, 0.24}

\title{\LARGE \bf Perching by hugging: an initial feasibility study}

\author{William Stewart$^{1}$, Mohammad Askari$^{1}$, Ma\"{i}k Guihard$^{1}$, and Dario Floreano$^{1}$
\thanks{This work was partially supported by the Swiss National Science Foundation through the National Centre of Competence in Research (NCCR). This work was also partially funded by the European Union’s Horizon 2020 research and innovation programme under grant agreement ID: 871479 AERIAL-CORE \emph{(Corresponding Author: William Stewart)}}
\thanks{$^{1}$The authors are with the Laboratory of Intelligent Systems, Ecole Polytechnique Federale de Lausanne, Lausanne CH1015, Switzerland e-mail:
        {\tt\small william.stewart@epfl.ch}}%
}

\begin{document}

\maketitle
\thispagestyle{empty}
\pagestyle{empty}

\begin{abstract}

Current UAVs capable of perching require added structure and mechanisms to accomplish this. These take the form of hooks, claws, needles, etc which add weight and usually drag. We propose in this paper the dual use of structures already on the vehicle to enable perching, thus reducing the weight and drag cost associated with perching UAVs. We propose a wing design capable of passively wrapping around a vertical pole to perch. We experimentally investigate the feasibility of the design, presenting results on minimum required perching speeds as well as the effect of weight distribution on the success rate of the wing wrapping. Finally, we comment on design requirements for holding onto the pole based on our findings.

\end{abstract}

\section{INTRODUCTION}

UAV (Unmanned Aerial Vehicle) perching is a topic of great interest to researchers at the moment. Increasingly precise sense and control systems are making it increasingly feasible to implement perching. As these control systems mature, physical mechanisms to perch with are becoming more common. Solutions for multicopters abound, including passive, rigid, avian-inspired claws used to perch on branches and bars \cite{passive_perch}, compliant avian-inspired claws \cite{hanging_branches_conference}, and microspines used to hold onto the sides of walls \cite{SCAMP-Overview}. However, there are fewer fixed-wing UAV solutions. One of the simplest proposals is the use of hooks to hang on a wire \cite{old_exp}. The hooks are small and lightweight, keeping drag and weight to a minimum, but are only effective when perching on a wire. Kovac et al. demonstrated a passive system for fixed-wing perching on walls that consisted of spring-loaded needles that are driven into the wall when the aircraft hits it straight on \cite{Needles_Kovac}. Although this system is limited to very lightweight UAVs, it was the first system to consider a crash-landing approach to perching. This has the considerable advantage of avoiding a dangerous, but commonly used, pitch-up maneuver to bleed off speed before perching. 

All of these solutions require some additional hardware, be it arms, claws, hooks, needles, etc. This can result in significant added mass and oftentimes added drag to the UAV, which offsets some of the benefits gained by perching. A better approach would be to take advantage of structure that is already on the vehicle and use that to perch. The natural choice for winged UAVs is the wings. We propose to add morphing capabilities to the wings of a UAV to enable them to wrap their wings around long linear infrastructure such as vertical poles and bars. 

\begin{figure}[tbp]
\centering
\begin{center}
\includegraphics[trim = 0mm 0mm 0mm 0mm, clip, width=0.5\textwidth]{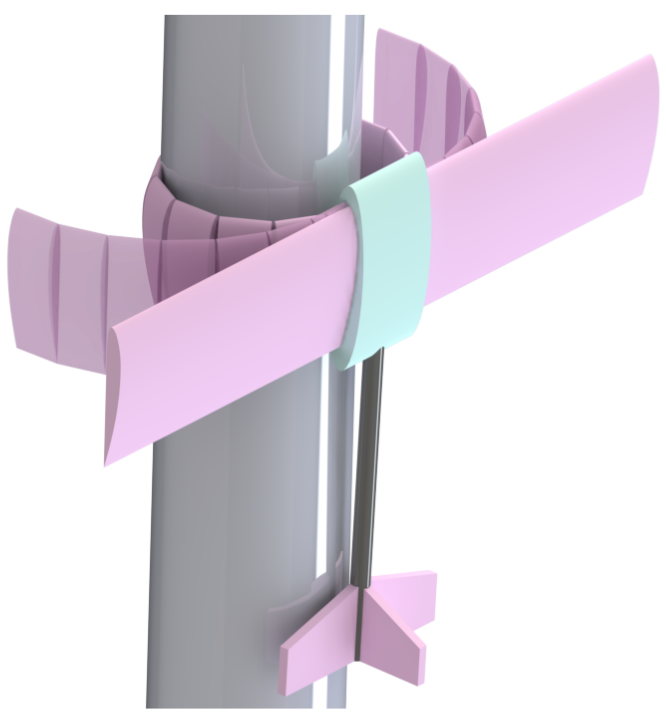}   
\caption{CAD rendering of a UAV using wing wrapping to perch.}
\label{perching_strat_diagram}
\end{center}
\end{figure}

None of the current perching solutions have demonstrated an ability to perch on a vertical pole. Yet, these kinds of structures are ubiquitous throughout the world and examples include man-made structures such as street lights, telephone poles, and building scaffolding. Perching on a pole would also be helpful in applications such as nature conservation or natural resources monitoring due to the abundance of these structures in nature, such as tree trunks. Thus, there are a lot of potential perching locations that wing-wrapping could be used and are not possible with current perching strategies.

\begin{figure*}[!ht]
\centering
\begin{center}
\includegraphics[trim = 0mm 0mm 0mm 0mm, clip, width=1.0\textwidth]{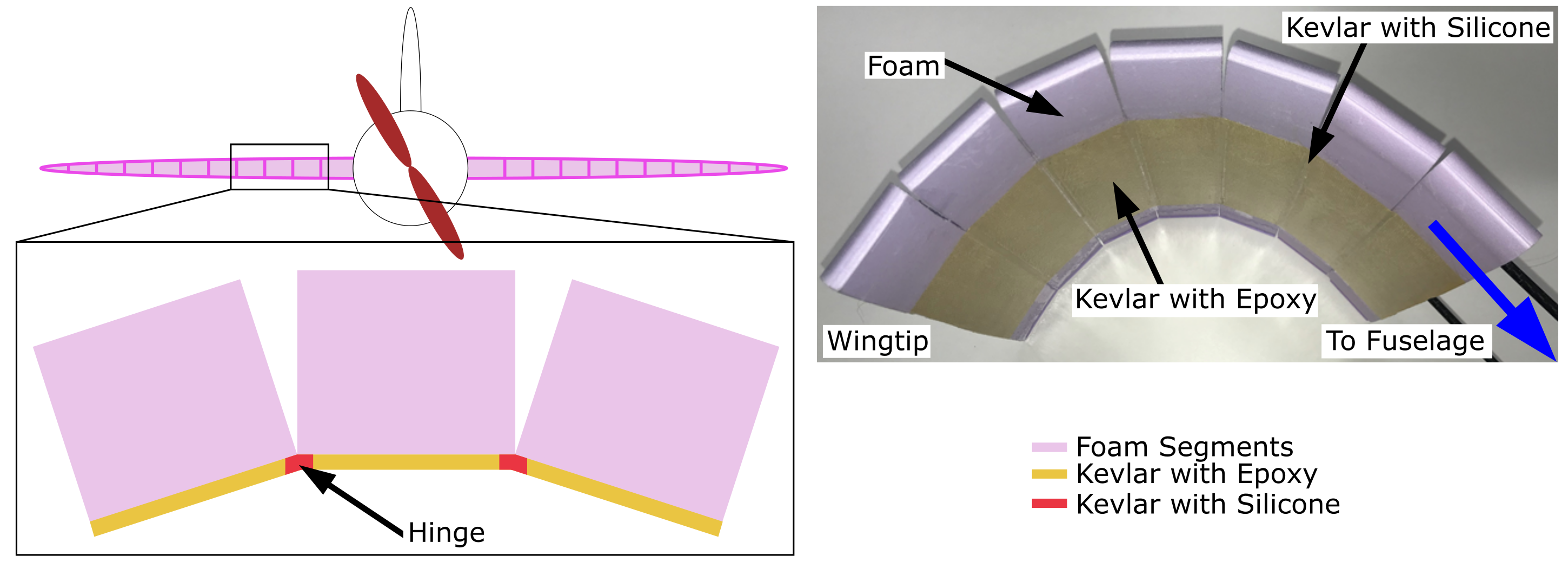}    
\caption{Diagram of morphing wing structure (left) and photograph of one wing (right).}
\label{show_our_design}
\end{center}
\end{figure*}

There are two rotational directions that the wings could use to wrap around a pole. The first is in the sweeping direction, where the wingtips move about an axis parallel to the vertical axis and the second is in the dihedral direction where the wingtips move about an axis parallel to the longitudinal axis. Due to the lower area moment of inertia of the wings in the longitudinal plane, we first investigated morphing the wings in this direction. In fact, many mechanisms already exist for morphing wings in this direction as a result of studies on flapping wing UAVs \cite{lentink_flap, bat_flap}. 

\begin{figure}[tbp]
\centering
\begin{center}
\includegraphics[trim = 0mm 0mm 0mm 0mm, clip, width=0.5\textwidth]{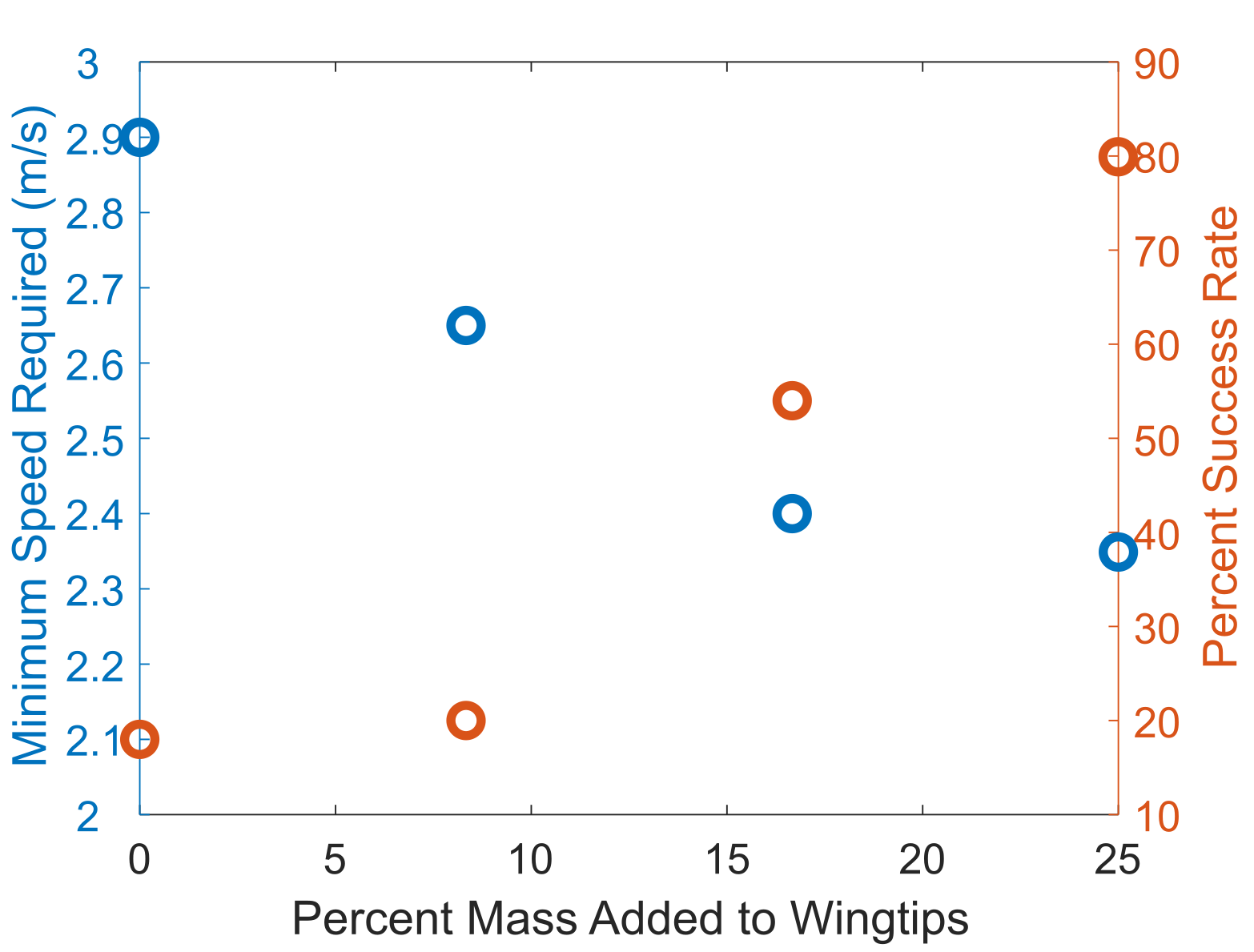}   
\caption{Experimental results. For a given added mass to the wing tips, is plotted both the percent success rate (red) and the minimum successful speed required to perch (blue).}
\label{data}
\end{center}
\end{figure}

We report here initial findings on the topic. We focused our work on understanding the feasibility of passive wing wrapping as a strategy for perching, and what limits it might have. The concept is that the UAV crashes into a vertical pole, and the inertia of the wingtips causes them to wrap around the pole at impact (Fig. \ref{perching_strat_diagram}). We designed and built a fuselage and wing structure capable of morphing about the longitudinal axis. Then, we used this mock-up aircraft to investigate whether at impact the wings would wrap around the pole, and what speed/weight distribution is best for wing wrapping. 

For this concept to function properly, the wing must be able to remain rigid in flight while at the point of impact, it must be able to wrap around the pole. In addition, the wing should be rugged enough to withstand the collision without major damage. The wing built for this study is a segmented morphing wing (Fig. \ref{show_our_design}). Each segment is connected to the next through a chord-wise hinge on the lower surface of the wing segments. The main structure of the segments is extruded polystyrene foam and they are connected by a single, spanwise sheet of Kevlar that is impregnated with epoxy to bond it with the lower surface of the segments. The hinge between the wing segments are created by impregnating the Kevlar with a band of silicone instead of epoxy. The softness of the silicone enables the Kevlar to bend and therefore the wings to curl. To ensure that the hinge is straight and can therefore rotate without binding, a flat-bottom airfoil, the clark-y, was used. Because the hinge is located on the lower surface of the wing, the lift generated in flight will keep the foam segments pressed against each other and not allow the wing to fold upwards. However, at impact, nothing is preventing the segments from rotating in the other direction, allowing them to be wrapped around the pole. The combination of durable Kevlar and soft foam has the advantage of being lightweight as well as able to absorb and damp impact forces, which can be high. The full vehicle is constructed by connecting the left and right wing halves to the fuselage with two carbon fiber rods.  

\begin{figure*}[!ht]
\centering
\begin{center}
\includegraphics[trim = 0mm 0mm 0mm 0mm, clip, width=1.0\textwidth]{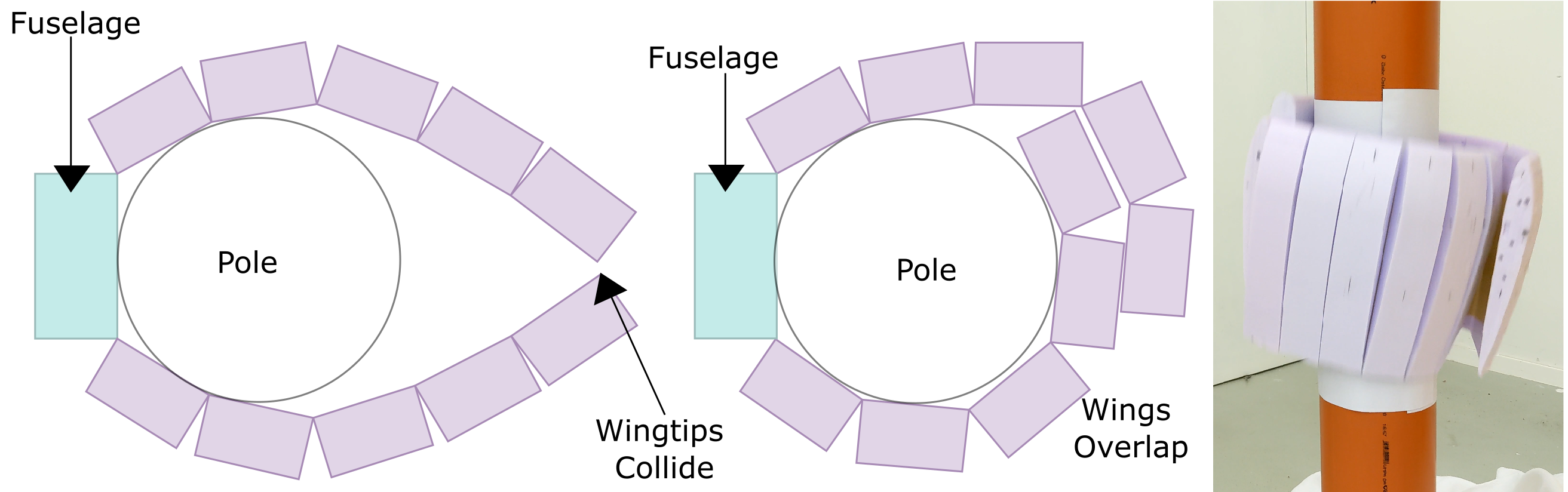}    
\caption{Diagram illustrating the case of when the wingtips collide (left) and wings overlap (middle). On the right is a photograph of the experiment showing the wings overlapping.}
\label{wingtips}
\end{center}
\end{figure*}

\section{EXPERIMENTS}

Initial experiments into wing-wrapping have considered only variations in speed and mass distribution. The objective of the experiment was to determine the minimum acceptable impact speed and how it changes with mass distribution. Because the concept is dependent on the inertia of the wingtips at impact, it follows that changing the impact speed or mass will affect the wings abilities to wrap around the pole. Higher speeds or wingtip masses should increase the success rate of unwrapping. Understanding these two effects can give an indication of how fast the UAV must fly to successfully wrap its wings around the pole and whether or not the strategy is feasible. 

The experiment consisted of hand tossing the vehicle onto a fixed pole with a diameter of 12cm. Position data from an OptiTrack motion capture system was used to measure impact speed. We adjusted the mass distribution by adding screws to the tips of the wings. We used four variations of mass, adding up to 25\% of the vehicles mass to the wingtips. Adding the mass both increased the success rate from less than 20\% to 80\% as well as reducing the minimum required speed at impact from 2.9m/s to 2.4m/s (Fig. \ref{data}). In a practical system, this could be implemented by adjusting the weight distribution of the vehicle to have more mass on the wingtips.  

In this experiment, we also noticed that sometimes the wings would wrap over one another and sometimes the wingtips would collide (Fig. \ref{wingtips}). In fact, over the 40 repetitions of the experiment,  40\% of the time the wings overlapped, and 60\% of the time, the wingtips collided. 

\section{DISCUSSION}

Our experiments showed that we were able to wrap the wings around the pole. We showed that a weight distribution that includes extra mass on the wing tips will increase the inertia of the wing tips, improving the success rate. However, we noticed that about 60\% of the time, the wingtips collided with one another as opposed to wrapping one over the other.

The next step in this line of research is to develop methods to keep the wings in the wrapped position. One possibility is using patches of velcro on the wings that stick together when the wings wrap over one another. Another is designing wingtips that can latch onto one another using an adhesive, gecko adhesion, or magnets. These solutions could even be combined into a single design that works regardless of whether the wingtips collide or wrap over one another. 

Once a mechanism that can keep the wings wrapped is in place, it will be necessary to address issues in staying on the pole. Throughout the experiments conducted here we noticed that after impact, the aircraft slid down the pole. To address this problem, we propose to use friction between the pole and aircraft. This in turn will require a normal force to \lq\lq hug\rq\rq the pole. In fact, if the normal force is applied throughout the entire wingspan, this could have the added benefit of helping keep the wings wrapped around the pole. 
\bibliography{IEEEabrv,EPFLBib}

\end{document}